\documentclass[letterpaper, 10 pt, conference]{ieeeconf}  

\IEEEoverridecommandlockouts                              

\usepackage{mathptmx} 
\usepackage{times} 
\usepackage{amssymb}  
\usepackage{graphicx}
\usepackage{amsmath, nccmath}
\usepackage{tabularx,ragged2e,booktabs,caption}
\usepackage{geometry}

\title{\LARGE \bf
	Identification Algorithm to Determine the Trajectory of Robots with Singularities}

\author{Hossein Sharifi, William C. Black}

\begin{document}

	\maketitle
	\thispagestyle{empty}
	\pagestyle{empty}

\begin{abstract}
Singularity is robot controls is an important problem. By identifying an appropriate trajectory for the robots, the singular situations can be avoided. In this paper an identification algorithm is proposed to control the robot such that it can change its direction to avoid the singularity situation. Base on the singular value decomposition, the proposed algorithm is developed for the non-redundant, single-rank robots. The proposed method is employed on a robot with six degrees of freedom, in order to identify its feasible trajectory. \\ 

\end{abstract} 

\section{Introduction}
Singularity occurrence in robot manipulators is considered as one of the challenging issues. In a singular situation, a robot may lose one or more degrees of freedom; meaning that the states in some directions will be zero [1-6]. The singularity may cause infinity speed in some joint. The directions which are affected by the singularity are known as the singular directions. In terms of a singularity in a joint, order reduction will happen in the Jacobian matrix. This order reduction will cause the matrix to be irreversible. Therefore, the control of the robot with irreversible Jacobian matrix will be extremely problematic [7-19]. 
\\ \indent
Several researchers have investigated the robot singularity in different ways. In [7], the robot joint velocities in singular situation have been surveyed; however, the authors only considered the perpendicular directions when a singularity happens, to determine the feasible paths. In another study [8], the authors have studied several control algorithms to control the robot's singularity; the methods are based on the inverse kinematic approach. One of the most important problems in robot singularity is that, the robot tends to move toward the singular direction once a singularity happens. To solve this issue, in this paper, we will analyze the robot trajectory in singular situation, and we will determine a feasible trajectory to avoid singularity through a proposed identification algorithm. In the proposed approach, through the identification algorithm, we will identify the possible paths once the singularity happens. Then, the robot will be directed to the trajectories that help it to avoid the singularity. The developed algorithm in this paper is based on the singular value decomposition method. The feasible paths in this method are the ones that result in finite outputs for the finite inputs (guarantees stability). Authors in [9] have proposed a similar approach based on strong equivalency for an electric network. They have identified the system through their proposed identification algorithm, such that they can guarantee avoiding singularity in the system. 
\\ \indent
In the rest of the paper, you will read section II which explains the mathematical formulation of the singular model, and the singular velocity model for the robot. Section III contains the identification of feasible paths in singularity. The proposed method is implemented on a robot with six degrees of freedom in section IV. Eventually, section V is the conclusion section. 
 \\
\section{Mathematical Model Definition}
As stated in the previous section, singularity results in irreversible Jacobian matrix, due to the matrix order reduction. The relationship between the velocity of joints and velocity of the end-points is represented as Eq. \ref{eq:velocity} [10-17].
In the following equation, $J$ is the Jacobian matrix, $\dot{q}$ represents the velocity of joints, and $\dot{X}$ denotes the velocity of the end-points, $n$ represents the number of joints, and $m$ is the number of degrees of freedom.  \\ \indent

\begin{equation} 
\begin{aligned}[b] 
\dot{X}&=J \times \dot{q}
\end{aligned}
\label{eq:velocity}
\end{equation} \\

Based on the singular value decomposition method, the Jacobian matrix can be decomposed into the multiply of three matrices as Eq. \ref{eq:J}. Moreover, the $V=[v_1,v_2, ..., v_n]$ matrix is a perpendicular matrix which is the multiply of matrices $J$ and $J^T$. We want to mentio that, the parameters can be tuned through the adaptive or fuzzy algorithms as in [15] and [17]. \\ \indent

\begin{equation} 
\begin{aligned}[b] 
J&=U \times \sum V^T
\end{aligned}
\label{eq:J}
\end{equation}  \\

Furthermore, $U$ contains the eigen vectors of $J \times J^T$.  Matrix $\sum$ can be defined also as Eq. \ref{eq:sum}. \\ \indent

\begin{equation} 
\begin{aligned}[b] 
\sum &=
\begin{bmatrix} 
\hat{\sum}_[r \times r] & 0_{[r \times (n-r)]}\\
0_{[(m-r) \times r]} & 0_{[(m-r) \times (n-r)]}\\
\end{bmatrix}
\end{aligned}
\label{eq:sum}
\end{equation} \\

In above equation, $r$ denotes the rank of the Jacobian matrix, and the $0$s are the zero matrices with the associated dimensions. The diagonal elements of $\hat{\sum}$ are calculated from the squared root values of the eigen values of matrix $J \times J^T$ as Eq. \ref{eq:r}. 

\begin{equation} 
\begin{aligned}[b] 
\sigma_1 \geq \sigma_2 \geq ... \geq \sigma_m
\end{aligned}
\label{eq:r}
\end{equation} \\

From the eigen value decomposition method, inverse of the Jacobian matrix is calculated in Eq. \ref{eq:inverse}. \\

\begin{equation} 
\begin{aligned}[b] 
J^* &= V \times \sum ^ * \times U^T, \\
\sum ^ * &=
\begin{bmatrix} 
\hat{\sum ^ *} & 0_{[r \times (n-r)]}\\
0_{[(n-r) \times r]} & 0_{[(n-r) \times (m-r)]}\\
\end{bmatrix}
\end{aligned}
\label{eq:inverse}
\end{equation} \\

The $\hat{\sum ^ *}$ matrix is calculated from Eq. \ref{eq:sigma_hat}. \\

\begin{equation} 
\begin{aligned}[b] 
\sum ^ * = diag (\frac{1}{\sigma_1}, \frac{1}{\sigma_2}, ..., \frac{1}{\sigma_r})
\end{aligned}
\label{eq:sigma_hat}
\end{equation} \\

In a singularity case, $\sigma_1, \sigma_2, ..., \sigma_m = 0$. Therefore, the last columns of $U$ matrix (from $u_{r+1}$ to $u_m$) are zero. The singular direction here is determined based on the direction of $u_m$. Thus, we have the following equation. \\

\begin{equation} 
\begin{aligned}[b] 
\dot{d}={u_m}^T \times \dot{X}
\end{aligned}
\label{eq:direction}
\end{equation} \\

In the above equation, $\dot{Y}$ denotes the $m-1$ non-singular vectors, and $\dot{d}$ is the velocity vector in a singular direction. Thus, Eq.  \ref{eq:velocity} can be rephrased as Eq. \ref{eq:rephrase}.  \\

\begin{equation} 
\begin{aligned}[b] 
\dot{Y}&=K \times \dot{q}  \\
\dot{d}&=L \times \dot{q} 
\end{aligned}
\label{eq:rephrase}
\end{equation} \\

In the above equation, $K$ and $L$ are the sub-matrices of $U^T \times J$. In a singular situation, $\dot{d}$ is zero, and $L$ is also zero. We should mention that the velocity matrix $\dot{q}$ is non-zero. \\

\section{Identification of Feasible Paths}
 By derivation of Eq. \ref{eq:rephrase}, and knowing that $L=0$, $\dot{d}=0$, the singular acceleration is represented as: \\
 
 \begin{equation} 
\begin{aligned}[b] 
\ddot{d}=\frac{dL}{dt} \dot{q} = (\dot{q}^T \times \theta \times L^T) \times \dot{q}
\end{aligned}
\label{eq:Ddoubledot}
\end{equation} \\
 \begin{equation} 
\begin{aligned}[b] 
\theta=
\begin{bmatrix} 
\frac{dl_1}{dq_1} & \frac{dl_2}{dq_1} & ... & \frac{dl_n}{dq_1} \\
\frac{dl_1}{dq_2} & \frac{dl_2}{dq_2} & ... & \frac{dl_n}{dq_2} \\
... & ... & ... & ...  \\
\frac{dl_1}{dq_n} & \frac{dl_2}{dq_n} & ... & \frac{dl_n}{dq_n}  
\end{bmatrix}
\end{aligned}
\label{eq:teta}
\end{equation} \\

According to $\dot{Y}$ matrix in Eq. \ref{eq:rephrase}, the $K$ matrix is a full-rank matrix, with $rank = m-1$. It can be defined as Eq. \ref{eq:K}.  $K_p$ in this equation is a reversible square matrix, with $m-1$ number of rows and columns, and $K_s$ is a matrix with dimension $(m-1) \times [(n-m+1)]$. \\

\begin{equation} 
\begin{aligned}[b] 
K=[K_p, \,\,\, K_s] 
\end{aligned}
\label{eq:K}
\end{equation} \\

The $\dot{q}$ matrix also can be decomposed into $\dot{q}_p$ and $\dot{q}_s$ with $(m-1)$, and $(n-m+1)$ dimensions respectively. Therefore, we can rewrite Eq. \ref{eq:rephrase} as Eq. \ref{eq:reqrite}.  \\

\begin{equation} 
\begin{aligned}[b] 
\dot{Y}=K_p \times \dot{q_p} + K_s \times \dot{q_s}
\end{aligned}
\label{eq:reqrite}
\end{equation} \\

Also, from the previous equations, we have $\dot{q}=M \times \dot{Y} + N \times \dot{q_s}$. Eventually, the equation for the acceleration in the singular direction is attained as Eq. \ref{eq:acc_final}. Matrices $A$, $B$, and $C$ are defined as Eq. \ref{eq:ABC}. \\

\begin{equation} 
\begin{aligned}[b] 
\ddot{d}=\dot{q_s}^T \times A \times \dot{q_s} + B \times \dot{q_s} + C
\end{aligned}
\label{eq:acc_final}
\end{equation} \\
\begin{equation} 
\begin{aligned}[b] 
A &= N \times \theta \times L \times N \\
B &= \dot{Y} \times M^T (\theta \times L^T + \theta \times L) \times N \\
C &= \dot{Y}^T \times M^T \times \theta \times L \times M \times \dot{Y}
\end{aligned}
\label{eq:ABC}
\end{equation} \\

Having the above equations, we can analyze the singular situations with non-zero acceleration. here, we are analyzing the singularity in robots. Since the number of degrees of freedom in robots is equal to the number of end-points, the $\dot{q_s}$ vector will be an scalar, and Eq. \ref{eq:dotdot_d} is attained. \\

\begin{equation} 
\begin{aligned}[b] 
\ddot{d}=A \times \dot{q}_s ^2 + B \times \dot{q}_s + C
\end{aligned}
\label{eq:dotdot_d}
\end{equation} \\

In this paper, we assume that robot is in the beginning of its path and the singularity occurs. Therefore, $\dot{Y}(0)=0$. Based on Eq. \ref{eq:ABC}, matrices $B$ and $C$ will be zero too. Therefore, we will have the Equation below. If $A$ in Eq. \ref{eq:jadid} is non-zero, the acceleration sign will be the same as $A$ matrix sign. This means that the end-point will be the same sign as $A$. If $A$ matrix is positive, the end-point will be in the same direction of the singularity, and if it is negative, it will be in the adverse singularity direction. Therefore, we will pick the paths in which the acceleration sign in a singular situation is equal to matrix $A$ sign. If $A$ is zero, Eq. \ref{eq:jadid} will be as $\ddot{d}=0$, which means that the robot can not move with any acceleration value in the singular situation. As you could see, based on the matrix $A$ values, we can analyze the possible paths once a singularity happens. \\ 

\begin{equation} 
\begin{aligned}[b] 
\ddot{d}=A \times \dot{q}_s ^ 2
\end{aligned}
\label{eq:jadid}
\end{equation} \\

\section{Proposed Path Identification Strategy on a Robot with Six Degrees of Freedom}
\begin{figure}[t]
    \begin{center}
                \centering
                \includegraphics[height=8cm, width=7cm]{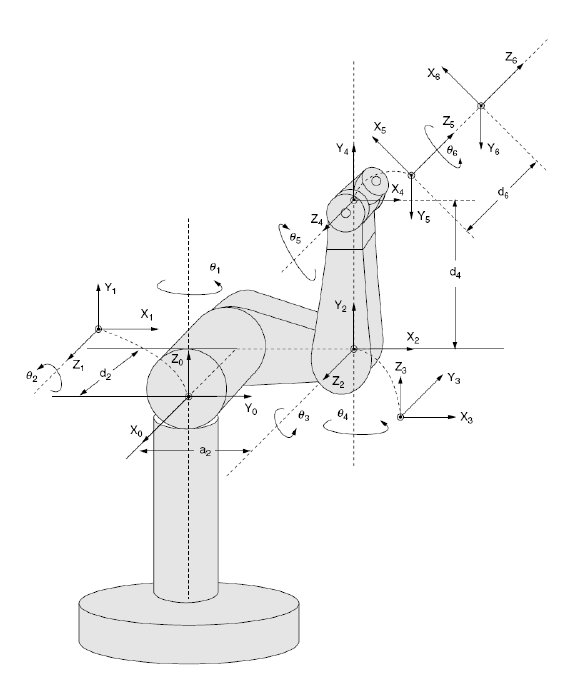}
                \caption{Robot picture with six degrees of freedom}
        \label{fig:robot}
    \end{center}
\end{figure}

In this section, our proposed path identification strategy is applied to a robot with six degrees of freedom. The robot is presented in Fig. \ref{fig:robot}, which has one vertical degree, and five circular degrees of freedom. The robot parameters are shown in Table \ref{tab:params}.

\begin{table}[t] 
        \caption{Robot parameters}
        \begin{tabular}{| m{2cm} | m{1cm} | m{1cm} | m{1cm} | m{1cm} |} 
            \hline
            {Joint} & {$\alpha _i$}  & {$a _i$} & {$d _i$} & {$\theta _i$}\\  \hline
            {First joint} & {$0$} & {$a_1$} & {$d_1$} & {$0$}  \\  \hline
            {Second joint} & {$0$} & {$a_2$} & {$0$} & {$\theta_2$} \\ \hline
            {Third joint} & {$90$} & {$0$} & {$0$} & {$\theta_3$} \\ \hline
            {Fourth joint} & {$-90$} & {$0$} & {$d_4$} & {$\theta_4$} \\ \hline
            {Fifth joint} & {$90$} & {$0$} & {$0$} & {$\theta_5$} \\ \hline
            {Sixth joint} & {$0$} & {$0$} & {$0$} & {$\theta_6$} \\ \hline
        \end{tabular}
        \label{tab:params}
\end{table}

 The Jacobian matrix for this robot will be as follows. \\
 
\begin{equation} 
\begin{aligned}[b] 
J=
\begin{bmatrix} 
J_{11} & J_{12} \\
J_{21} & J_{22}
\end{bmatrix}
\end{aligned}
\label{eq:robot_j}
\end{equation} \\

\begin{equation} 
\begin{aligned}[b] 
J_{11}=
\begin{bmatrix} 
0 & d_4 + a_2 \times S_3 & d_4 \\
1 & 0 & 0\\
0 & -a_2 \times C_3 & 0
\end{bmatrix}
\end{aligned}
\label{eq:j11}
\end{equation} \\

\begin{equation} 
\begin{aligned}[b] 
J_{12}=
\begin{bmatrix} 
0 & 0 & 0 \\
0 & 0 & 0\\
0 & 0 & 0
\end{bmatrix}
\end{aligned}
\label{eq:j12}
\end{equation} \\

\begin{equation} 
\begin{aligned}[b] 
J_{21}=
\begin{bmatrix} 
0 & 0 & 0 \\
0 & 1 & 1\\
0 & 0 & 0
\end{bmatrix}
\end{aligned}
\label{eq:j21}
\end{equation} \\

\begin{equation} 
\begin{aligned}[b] 
J_{22}=
\begin{bmatrix} 
0 & -S_4 & C_4 \times S_5 \\
0 & C_4 & S_4 \times S_5\\
1 & 0 & C_5
\end{bmatrix}
\end{aligned}
\label{eq:j22}
\end{equation} \\

Since, matrix $J_{12}$ is a zero matrix, the first three degrees of freedom are independent of the next two other degrees of freedom, and the solutions are independent of each other. The first three degrees of freedom eigen values would be as Eq. \ref{eq:solution}. \\

\begin{equation} 
\begin{aligned}[b] 
\sigma_1 &= 0.5 \times ({a_2}^2 + 2{d_4}^2 + 2a_2 d_4 S_3) + ({a_2}^2 + 2 {d_4}^2 + 2 a_2 d_4 S_3)^2  \\
&- 4(a_2 d_4 C_3)^2  \\
\sigma_2 &= 1 
\end{aligned}
\label{eq:solution}
\end{equation} \\

The singular values of the $\sigma_1$ and $\sigma_2$ are always positive and non-zero, therefore, the singular situation for the first three degrees of freedom happens only when $\sigma_3$ is zero; $\sigma_3 = k \pi + \frac{\pi}{2}$. The Jacobian matrix $J_{11}$ in a singular situation will be as the following equation. The $\epsilon$ value is 1 if $k$ is even, and -1, if $k$ is odd value. \\

\begin{equation} 
\begin{aligned}[b] 
J_{11}=
\begin{bmatrix} 
0 & d_4 + a_4 \times \epsilon & d_4  \\
1 & 0 & 0 \\
0 & 0 & 0
\end{bmatrix}
\end{aligned}
\label{eq:jac}
\end{equation} \\

The multiplication of $\theta$ with $L$ matrix would be as Eq. \ref{eq:multiply}. \\

\begin{equation} 
\begin{aligned}[b] 
\theta \times L = 
\begin{bmatrix} 
0 & 0 & 0 \\
0 & 0 & a_2 \times S_3 \\
0 & 0 & 0
\end{bmatrix}
\end{aligned}
\label{eq:multiply}
\end{equation} \\

And, the $K$ matrix would be as Eq. \ref{eq:K}, and $\dot{Y}$ is as \ref{eq:yydot}. \\

\begin{equation} 
\begin{aligned}[b] 
K = 
\begin{bmatrix} 
0 & d_4 + a_2 \times \epsilon & d_4 \\
1 & 0 & 0 
\end{bmatrix}
\end{aligned}
\label{eq:K}
\end{equation} \\
\begin{equation} 
\begin{aligned}[b] 
\dot{Y} = 
\begin{bmatrix} 
0 & d_4 \\
1 & 0 
\end{bmatrix}
\times 
\begin{bmatrix} 
\dot{q}_1 \\
\dot{q}_3 
\end{bmatrix}
+
\begin{bmatrix} 
d_4 + a_2 \times \epsilon \\
0 
\end{bmatrix}
\end{aligned}
\label{eq:yydot}
\end{equation} \\

And, $K_p$ and $K_s$ matrices would be as follows: \\

\begin{equation} 
\begin{aligned}[b] 
K_p = 
\begin{bmatrix} 
0 & d_4  \\
1 & 0 
\end{bmatrix}
\end{aligned}
\label{eq:Kp}
\end{equation} \\

\begin{equation} 
\begin{aligned}[b] 
K_s = 
\begin{bmatrix} 
d_4 + a_2 \times \epsilon  \\
0
\end{bmatrix}
\end{aligned}
\label{eq:Ks}
\end{equation} \\

Thus, the matrices $A$, $B$, and $C$ will be as follows: \\

\begin{equation} 
\begin{aligned}[b] 
A &= (a_2 \times \epsilon )^2 \times (- \frac{d_4 + a_2 \times \epsilon}{a_2 \times d_4 \times \epsilon}) \\
B &=C=0
\end{aligned}
\label{eq:ABC_jadid}
\end{equation} \\

Considering that $A$ is always non-zero, and the acceleration in a singular direction follows the sign of $A$ matrix, the possible paths are the ones that the acceleration sign is the same as $h$ parameter sign in Eq. \ref{eq:h}. 

\begin{equation} 
\begin{aligned}[b] 
h = - \frac{d_4 + a_2 \epsilon}{a_2 d_4 \epsilon} 
\end{aligned}
\label{eq:h}
\end{equation} \\

We conclude that if the $h$ sign is positive, the robot can avoid singularity in the direction of the singular eigen vector. If $h$ is negative, the robot will escape the singularity advers of the direction of the singular eigen vector. The mentioned diretions are considered as the limits for the robot. \\ \indent 
Also. from Eq. \ref{eq:ABC_jadid}, if parameters $d_4$ and $a_2$ are equal and $\epsilon$ is equal to -1, the $A$ matrix is zero. In this situation, there is no feasible path for the robot to avoid singularity. The same calculations are applied to the other/rest of degrees of freedom. \\

\section{Conclusion}
In this paper, feasible paths are detected to help the robot avoid singularities. The limit of this identification procedure are analyzed, and the proposed strategy is employed on a robot with six degrees of freedom. In this study, we assumed that the singularity happens when the robot is in the beginning of its path. The identification algorithm developed is based on the singular value decomposition method. The proposed strategy was very efficient and helpful in guiding the robot to avoid the singularity situation.


\end{document}